# A Health Monitoring System for Elder and Sick Persons

Ankit Chaudhary and Jagdish L. Raheja

*Abstract*—This paper discusses a vision based health monitoring system which would be very easy in use and deployment. Elder and sick people who are not able to talk or walk, they are dependent on other human beings for their daily needs and need continuous monitoring. The developed system provides facility to the sick or elder person to describe his or her need to their caretaker in lingual description by showing particular hand gesture with the developed system. This system uses fingertip detection technique for gesture extraction and artificial neural network for gesture classification and recognition. The system is able to work in different light conditions and can be connected to different devices to announce user's need on a distant location.

## I. INTRODUCTION

This system is targeted for the old or sick people, who are not able to express their feelings by words or they can't walk. Generally these people are in hospital or in home under continuous human monitoring. It is possible that the person who is monitoring is not nearby as he can also has some work and can go out from the room where he was monitoring. It is usually happen in hospitals where one hospital staff needs to take care of many patients which can be in different rooms. During this time if the sick or elder person wants to eat something or want to call someone for help, then they are unable to do that. Our developed system helps the elder/sick person to express their wish or need in predefined lingual description to a specified place. This place can be hospital staff room or living room in home, where generally people stay. The usage of system is very easy as the person is sick or not able to walk and he would not be able to do complex operations.

Our system is based on human to machine interaction, in which machine would be able to do action according to the predefined syntax of the gesture made by user. User can use his hand (either left or right) to make gesture with the developed system and its lingual description will be transferred to the specified place automatically. Only it is assumed that the users are able to move their hands, they are not fully disabled like in coma cases. This system has an interface which includes a small simple camera, where users have to show their hand in front of camera. In general when humans show their hands to others, they show palm side of hand. The same assumption we have in our system that palm should face the camera. This hand gesture would be interpreted by the system, whether it is valid gesture syntax or not. If the gesture is not included in the rules list, system will not take any action and will give a lingual message of wrong gesture to user. If it is valid then the system will describe it according to predefined data and user will be informed as action done. An initial draft on this work has been published as [1]. This system works on the principles of Computer Vision and it uses image processing for acquiring gesture and preprocessing. This system have 2D camera only and its cost is very much affordable. An algorithm flow of our system has been shown in the fig 1, which extract region of interest from the captured image and further detects hand on skin based detection. In gesture recognition color based methods are applicable because of their characteristic color footprint of human skin. The color footprint is usually more distinctive and less sensitive than in the standard (camera capture) RGB color space.

## II. BACKGROUND WORK

Many health monitoring commercial systems are available in the market at present. In few systems, it is necessary to wear it on their body and it senses or do some action if the person is able to walk. Also users are supposed to do press a button in one system. These systems are either embedded system or sensor based system. Generally the sick/elder people are so weak that they can't press any button and most of them can't walk. So our system is a better solution for them. In the literature, we found some works related to our system. As Mitra [2] defines gesture recognition a process where user made gesture and receiver recognized them. Many Researchers have done excellent work in this area. Ahn [3] have developed augmented interface table using infrared cameras for pervasive environment. Chaudhary [4] has described designing for intelligent systems in his work.

Most of color segmentation techniques rely on histogram matching or employ a simple look-up table approach [5]-[8] based on the training data for the skin and possibly it's surrounding areas. The major drawback of color based localization techniques is the variability of the skin color footprint in different lighting conditions. This frequently results in undetected skin regions or falsely detected non skin textures. The problem can be somewhat alleviated by considering only the regions of a certain size or at certain spatial position.

Another common solution to the problem is the use of restricted backgrounds and clothing like dark gloves or wearing a strip on wrist [9]-[12]. Wu [13] has been extracted hand region from the scene using segmentation techniques. Vezhnevets [14] describes many useful methods for skin modeling and detection. Skin color detection and boundary extraction are two important part of gesture extraction form the image. Gesture recognition is the phase in which the data analyzed from the visual images of gestures is recognized as a specific gesture. In this step we assume that gesture image has been extracted from the image captured and now it is target data for gesture recognition. Graph matching is widely used for object mapping in images, but it faces problems in dependency on segmentations [15]. The identification of a hand gesture can be done in many ways depending on the problem to be solved [12].

The interpretations of gestures require that dynamic or static configurations of human hand be measurable by the machine. First attempts to solve this problem resulted in mechanical devices called as glove-based devices that directly measure hand joint angles and spatial position [16]-[18]. In this system requires the user to wear a cumbersome device and carry a load of cables that connect the device to a computer. This hinders the ease and naturalness with which user can interact with computer controlled environment. Potentially, any awkwardness in using gloves and other devices can be overcome by video-based noncontact interaction techniques identifying gestures.

Most of the static models are meant to accurately describe the visual appearance of the user's hand as they appear to a human observer. To perform recognition of those gestures, some type of parameter clustering technique stemming from vector quantization (VQ) is usually used. Briefly, in vector quantization, an n-dimensional space is partitioned into convex sets using n-dimensional hyper planes, based on training examples and some metric for determining the nearest neighbor. Parameters of the model can be chosen especially to help with the recognition as in [19]-[20]. Structures like cylinders, spheres, ellipsoids and hyper-rectangles are often used to approximate the shape of gesture made like hand or finger joints, hand structure [21]. One approach is to build a classifier such as SVM as in [13], [22]. First of all in the captured images which are having hand positions, we have to define the hand structure.

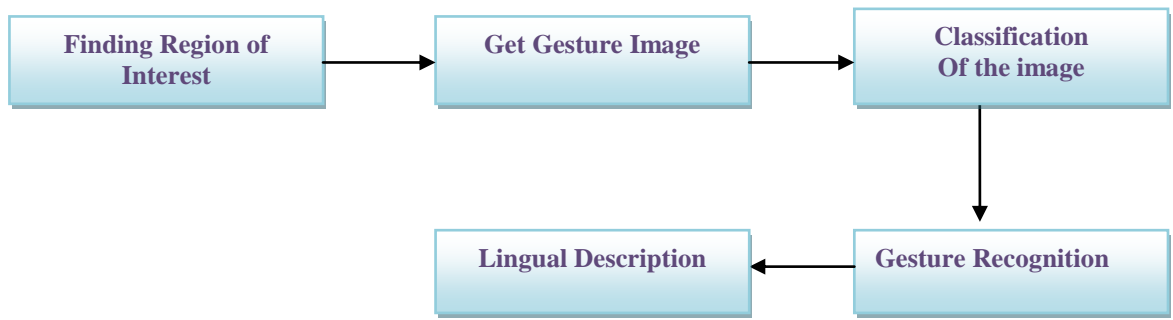

Fig. 1. System methodology

TABLE I: MODELING TECHNIQUES IN DIFFERENT APPLICATIONS

| S.No | Application | Gesture Modeling Technique |
|---|---|---|
| 1 | Fingerpaint [23] | Fingertip template |
| 2 | Finger pointer [24] | Heuristic detection of pointing system |
| 3 | Window manager [25] | Hand pose detection using neural networks |
| 4 | Automatic robot instruction [26] | Fingertip positions in 2D |

In complex systems such as GREFIT [27]-[28], the hand model is defined by a set of links and joints. Another technique is learning based methods which can be constructed any intelligent method. Few researchers use fingertip positions as parameters to construct hand image see table 1. This approach is based on the assumption that position of fingertips in the human hand relative to the palm, is almost always sufficient to differentiate a finite number of different gesture [16], [29]. Nguyen [12] uses fingertips recognition and detection based on applying a learning model, to reconstruct the hand model.

Fingers are the key points in hand gesture recognition modeling in the human hand model. The palm must be assumed to be rigid. Kerdvibulvech [30] uses a Gabor feature vector to extract the positions of a guitar player. Other approaches to get the open finger position information include (r, $\square$ ) and B-Spline curves or using the curvature information by considering flow between arcs and fingers in the hand model. A different way to detect fingertips is to use pattern matching techniques as templates and can be enhanced by using additional image feature like contours [31].

### III. PLAN OF WORK

A major motivation for this work on gesture recognition is the potential to use hand gestures for the general applications, aiming for the natural interaction between the human and various computer controlled displays. The current research in this area is at high pace but still further theoretical as well as computational advancements are needed before gestures can be widely used for human computer interaction in natural environment. The problem of accurate recognition of gestures which use model parameters that cluster in non-convex sets can also be solved by selecting nonlinear clustering schemes. Artificial neural networks are one such option, although their use for gesture recognition has not been fully explored [25]. The detection of human fingertips in the human hand structure is an important issue in most hand model studies and in some gesture recognition systems [12]. It is also clear form literature survey that the work done in this area is not sufficient and requires and new keen investigations in this direction. This area has many possibilities in the field of Computer Vision and Human machine interaction. Gesture recognition could be based on probability, if the background is not static or image has other same kind of objects that would also lead to imprecise results.

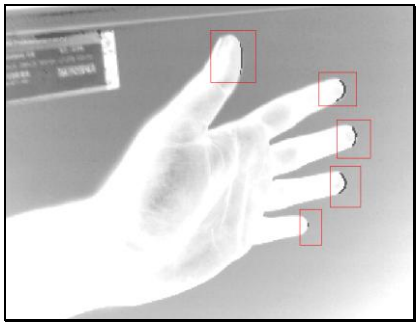

Fig. 2. Results of fingertip detection in image frame

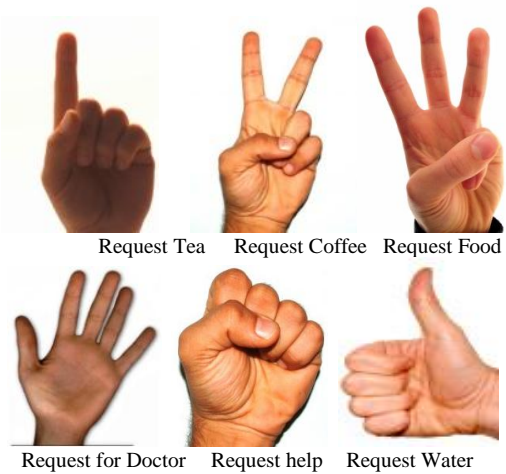

Fig. 3. Predefined gestures and their lingual discriptions

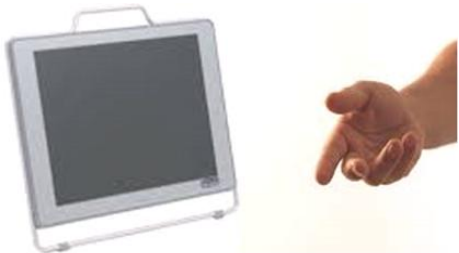

Fig. 4. Ease of use

Generally there are three phases in gesture recognition process - image capture, gesture extraction and gesture recognition. These phases include several issues such as algorithm design, processing speed, system architecture and video interface [15]. Cameras can detect people and recognize their activities in an application environment such as room, a plane, a car or a security checkpoint. The results of these cameras can be analyzed and used to control the operation of devices in these application environments. For the simplicity we focused only on 2D static gesture recognition. We have captured the image using a high resolution camera and after preprocessing, fingertip detection results are shown in Fig. 2. For light invariant system, we used orientation histograms to match test gesture with the stored database images.

The system use only 6 predefined gestures as shown in fig 3. They are configurable and even their descriptions also can be changed. Addition of new gesture is allowed at any time according to user requirement. These gestures are generic to the elder/sick need. User only has to show one of these gestures to the system and information would be announced to the staff or to family member depend on the location. The usage of system is very easy as shown in Fig. 4.

## IV. CONCLUSION

Hand Gesture Recognition is very easy and natural way to interact with the machines where no training for the user is required. So future of this technique is very bright in systems for handicap and disabled persons. Here we presented a system for the elder/sick person, assuming they are on bed and can't express them. So this system would help them in expressing their needs to others. This system has been presented at [32]. This technique can be applied in the field of disaster management so that lives of more people could be saved. In the damage situations i.e. in mining where, workers are always in danger zone. If mine collapse then workers would be under the pile of dust or coal and if he is alive, still he is not able to communicate to other people or to rescue team, for saving his life. This system will be placed on the body of each miner or persons staying or working in such dangerous places and will put the transceiver on the nearby wall. So in any case of disaster, if the user is in danger and can't go to ground level, he can show predefined gesture syntax to system that will interpret it and send a signal to transceiver nearby and it will forward the signal further to the rescue team in the control room. In this scenario an advanced version of this system is required, which would be GPS enabled to detect the exact location of the user in the mine. It will work in the case of suffocation, gas poisoning or fire hazards where the person is not able to shout or to tell even a single word.

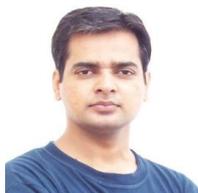

**Ankit Chaudhary** received his Master of Engineering Degree in Computer Science and Engineering from Birla Institute of Technology & Science, Pilani, India and Ph.D. in Computer Vision, from CEERI-BITS Pilani, India. His areas of research interest are Computer Vision, Artificial Intelligence and Intelligent Systems.

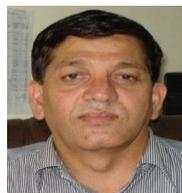

**Jagdish L. Raheja** Received his M.Tech from Indian Institute of Technology, Kharagpur, INDIA and Ph.D. from Technical University of Munich, Germany. Currently he is Director, Machine Vision lab at Central Electronics Engineering Research Institute (CEERI), Pilani, India. His areas of research interest are Digital Image Processing, Embedded Systems and Human Computer Interface.